\pdfoutput=1

\documentclass[11pt]{article}

\usepackage[table]{xcolor}
\usepackage{acl}

\usepackage{times}
\usepackage{latexsym}

\usepackage[T1]{fontenc}

\usepackage[utf8]{inputenc}

\usepackage{microtype}
\usepackage{booktabs}
\usepackage{inconsolata}

\usepackage{amsmath}
\usepackage{graphicx}
\usepackage{caption}

\usepackage{adjustbox}
\usepackage{multirow}
\usepackage{enumitem}
\usepackage{geometry}
\usepackage{amsmath}
\usepackage{booktabs}

\title{Mixed Distillation Helps Smaller Language Model Better Reasoning}

\author{
    Chenglin Li$^{1*}$,
    Qianglong Chen$^{1}\thanks{$\quad$ Equal Contribution.}$, 
    Liangyue Li$^{3}$,
    Caiyu Wang$^{2}$
    \\
    \textbf{Yicheng Li}$^{1}$,
    \textbf{Zulong Chen}$^{3}$,
    \textbf{Yin Zhang}$^{1}$
    \\
    $^{1}$Zhejiang University, Hangzhou, China \\
    $^{2}$Dalian Medical University, Dalian, China \\
    $^{3}$Alibaba, Hangzhou, China \\
    \texttt{\{22351307,chenqianglong,22321334, zhangyin98\}@zju.edu.cn } \\
    \texttt{\{jackiet99,wangcaiyu95,chenzulong198867\}@gmail.com}
}

\begin{document}
\maketitle
\begin{abstract}
While large language models (LLMs) have demonstrated exceptional performance in recent natural language processing (NLP) tasks, their deployment poses substantial challenges due to high computational and memory demands in real-world applications. Recent studies have focused on enhancing smaller models through knowledge distillation from LLMs, yielding promising results. 
However, these models often struggle to match the performance of LLMs, especially in tasks that require reasoning. In this work, we introduce \textbf{Mixed Distillation} (MD) framework, which capitalizes on the strengths of Program of Thought (PoT) and Chain of Thought (CoT) capabilities within LLMs, combining multiple prompting techniques and distilling these capabilities into smaller models. Our experimental results show that MD significantly enhances the single-path and multi-path reasoning ability of smaller models in various tasks. In terms of accuracy and generality of reasoning tasks, the model generated by it exceeds the comprehensive performance of two individually distilled models. Notably, LLaMA2-7B and CodeLlama-7B using MD achieved remarkable improvements of \(84.5\%\) and \(85.5\%\), respectively, outperforming GPT-3.5-Turbo by \(2.5\%\) and \(3.5\%\), on the SVAMP benchmark.

\end{abstract}
\begin{figure}
    \centering
    \includegraphics[width=1.0\linewidth]{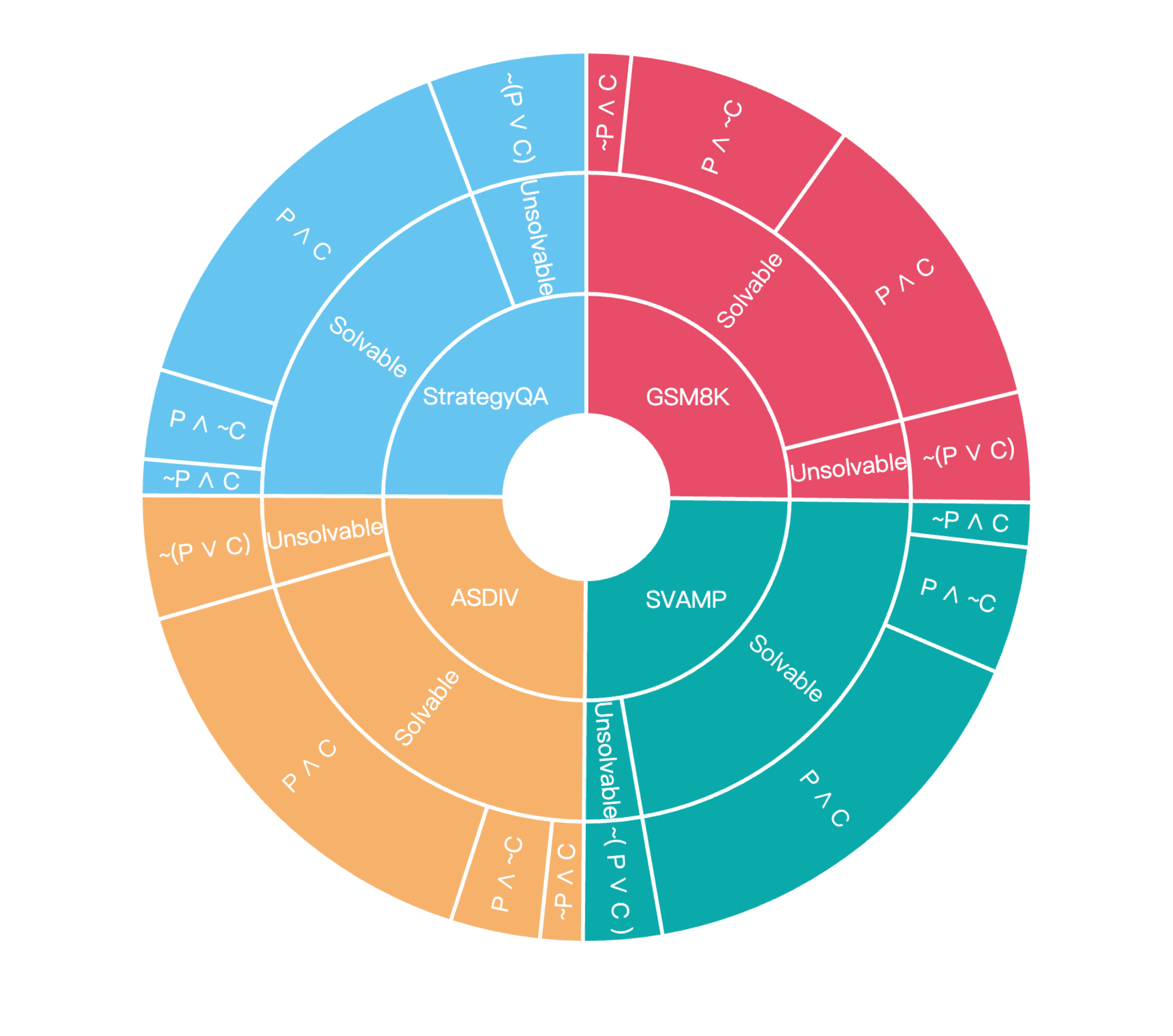}
    \vspace{-10mm}
    \caption{Evaluate the problem solving ability of GPT-3.5-Turbo across datasets. $\sim P \land C$ denotes CoT's unique solutions and $P \land C$ represents shared solvability. $\sim(P \lor C)$ indicates the remaining unsolved challenges.}
    \label{statics}
\end{figure}

\begin{figure*}
    \centering
    \includegraphics[width=1.0\linewidth]{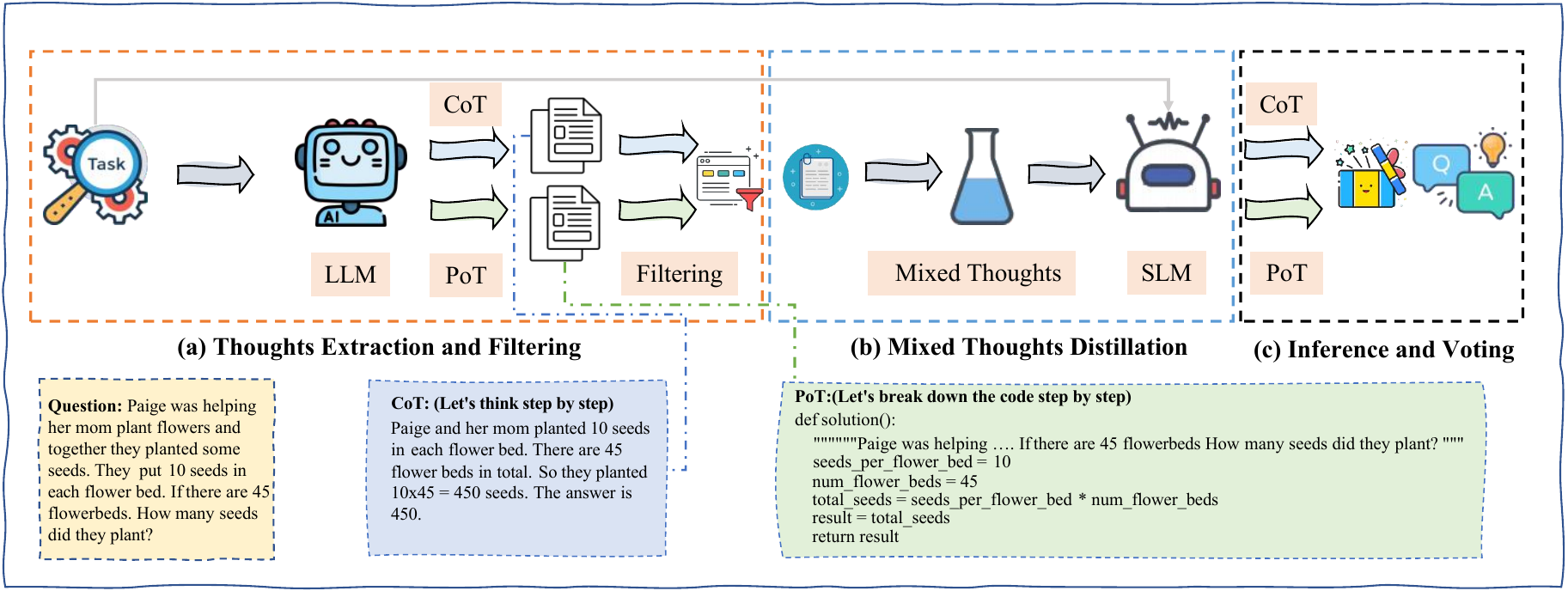}
    \caption{Overview of Mixed Distillation framework: extracting and distilling Chain of Thought and Program of Thought from large language models (LLM) to task-specific smaller language models (SLM).}
    \label{framework}
\end{figure*}

\section{Introduction}

Recent LLMs have made great progress in the field of natural language processing (NLP)
~\cite{brown2020language,kojima2022large,wei2022chain,ouyang2022training,touvron2023llama}, which exhibit the capability to explain their predictions by generating intermediate reasoning steps  ~\cite{kojima2022large,wei2022chain,thoppilan2022lamda,hoffmann2022training,zheng2022alpa}. 
Some works leverage these intermediate reasoning steps to improve the few-shot~\cite{brown2020language} or zero-shot~\cite{kojima2022large} setting performance of LLMs, while other works consider these reasoning steps as additional fine-tuning datasets for self-improving~\cite{huang2022large,zelikman2022star}. Meanwhile, LLM has become a powerful tool~\cite{wang2023survey,zhao2023survey,thirunavukarasu2023large,min2023recent} after pre-training large-scale data containing codes~\cite{ouyang2022training,touvron2023llama}. 

However, the deployment of these advanced LLMs in real-world applications presents significant costs and computational demands~\cite{kaplan2020scaling,sorscher2022beyond}. To address these challenges, distilling the capabilities of LLMs emerges as a resource-friendly and effective strategy, especially for specific tasks. Through knowledge distillation, prior works can efficiently distill the basic abilities of LLMs into smaller models~\cite{tang2019distilling,wang2021want,smith2022language,arora2022ask}. However, they mainly focus on distilling the CoT capability of LLMs into smaller models and ignore the critical PoT as a supervisory signal. Meanwhile, compared with LLMs, smaller models struggle with generating effective intermediate steps due to limited knowledge, which makes their reasoning ability challenging~\cite{valmeekam2022large,huang2022towards,chu2023survey}.

On the other hand, LLMs' PoT and CoT capabilities can solve specific problem subsets.
As shown in Figure \ref{statics}, LLM can address the majority of solvable problems (including $P \land C$, $\sim P \land C$, $P \land \sim C$). Among them, a large part of the problems can be solved by both PoT and CoT ($P \land C$). However, each ability still has its advantages. Specifically, only about 6\% of the problems can be solved exclusively via CoT ($\sim P \land C$) across tasks. In contrast, PoT exclusively addresses 31.98\% of problems on the GSM8K dataset, and this figure exceeds 10\% across other datasets ($P \land \sim C$). More details are shown in Appendix~\ref{casellm}. In addition, \cite{zhang2021survey,wei2021finetuned,longpre2023flan} have proved that multi-task learning enhances model performance by involving various knowledge domains. Therefore, distilling both reasoning capabilities is likely to yield the best outcomes.

In light of these insights, we propose a novel distillation framework, Mixed Distillation, as shown in Figure ~\ref{framework}. 

\begin{figure}
    \centering
     \includegraphics[width=1.0\linewidth]{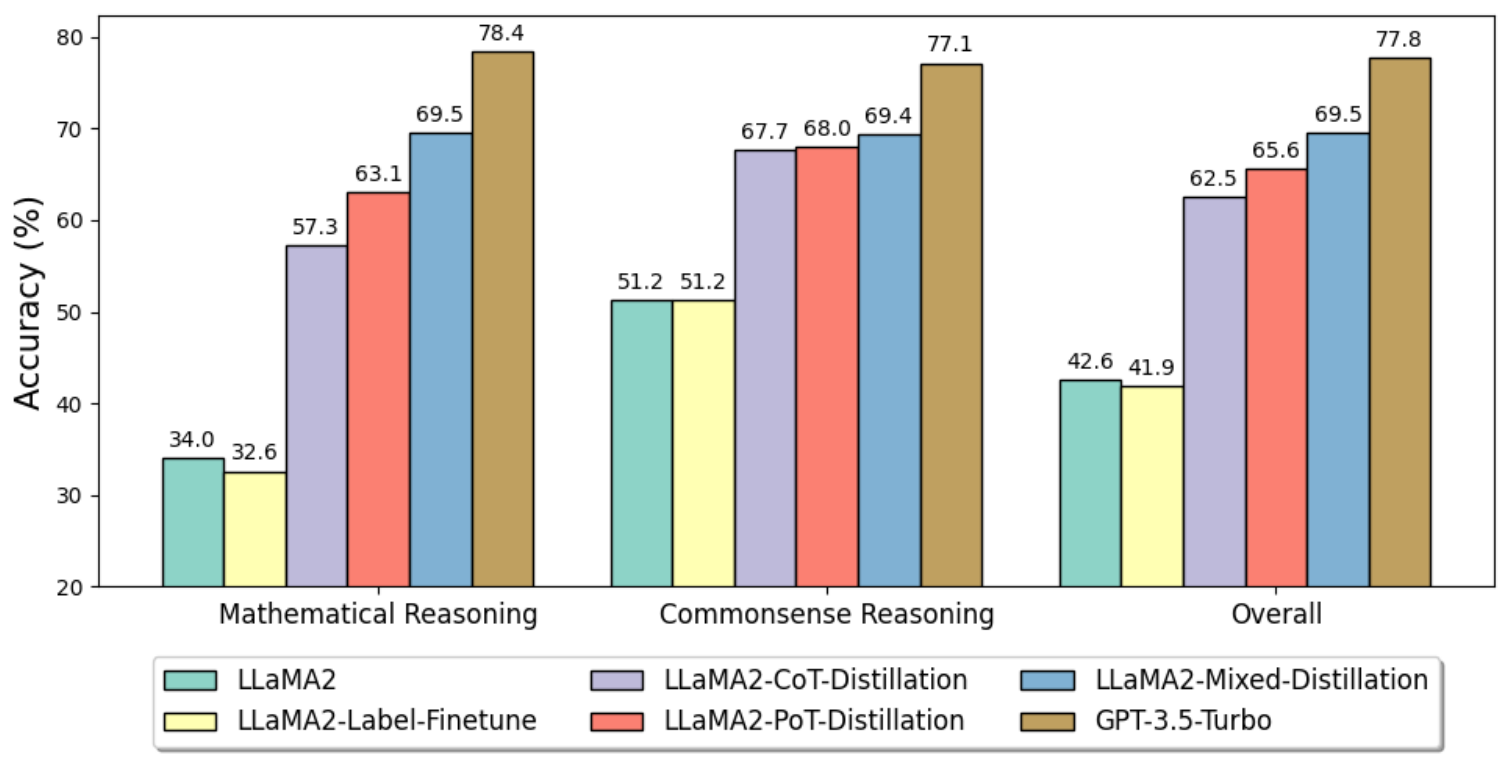}
    \caption{Performance of different methods across reasoning domains based LLaMA2-7B.}
    \label{models_accuracy}
\end{figure}

In contrast to Single-Path Distillation which relies on a sole supervisory signal, Mixed Distillation (MD) utilizes CoT and PoT as combined supervisory signals in the multi-task learning framework for smaller models. Additionally, the smaller models adopt multi-path reasoning by generating CoT and PoT during inference.
Our experimental results show that MD enhances the reasoning ability of smaller models. As shown in Figure \ref{models_accuracy}, the overall performance of the smaller model in the task is better than that of the model using single path distillation after mixed distillation. Notably, on SVAMP, LLaMA2-7B achieves a significant 15\% improvement over CoT Distillation, and T5-Large records an impressive 43.5\% increase compared to CoT Distillation.
The contributions of our work are as follows:
\begin{itemize}

\item We emphasize that PoT is a novel and effective supervisory signal for model distillation, which can enhance the reasoning ability of smaller models in specific tasks.
\item To enhance the smaller models' reasoning including single-path and multi-path reasoning, we propose a novel framework, Mixed Distillation, which combines multiple prompting techniques, marking a significant advancement in model distillation.
\item We conducted a series of experiments to highlight the efficacy of Mixed Distillation, revealing notable improvements in reasoning capabilities across smaller models, across datasets of varying scales and even extending to out-of-distribution training data.
\end{itemize}


\section{Related Work} 
\paragraph{Multiple Thoughts Prompting Techniques in LLM} 
Recent work\cite{wei2022chain,chu2023survey,gao2023pal,chen2022program,hu2023code,imani2023mathprompter}, focusing on eliciting the thought process of LLMs, has validated its effectiveness in the domain of reasoning, such as SVAMP~\cite{patel2021nlp}, GSM8K~\cite{cobbe2021training}, ASDIV~\cite{miao2021diverse} and StrategyQA\cite{geva2021did}.
CoT \cite{wei2022chain} enhances reasoning by prompting LLMs to generate intermediate natural language thought processes.
PoT~\cite{gao2023pal,chen2022program} stimulates LLM's reasoning ability by prompting them to generate intermediate code that can be executed by the python executor.  ~\citet{zhang2023natural,li2023chain} illustrates the adaptability of seamlessly combining natural language reasoning and program synthesis within prompt-based learning to effectively solve reasoning tasks. 
~\citet{imani2023mathprompter,yue2023mammoth} improve the mathematical reasoning ability of LLMs combining PoT and CoT by designing prompts. ~\citet{yue2023large} develop a cost-effective approach by combining PoT and CoT. However, previous work provided limited insights into PoT and CoT in improving smaller models, and our MD fills the gap in exploring how small models can utilize mixed thoughts to enhance the performance in reasoning.

\paragraph{Knowledge Distillation from LLMs}
Knowledge distillation~\cite{buciluǎ2006model, ba2014deep,hinton2015distilling,beyer2022knowledge,fu2023specializing} has proved its effectiveness in improving smaller models. Some works~\cite{li2023symbolic,hsieh2023distilling,wang2023democratizing,fu2023specializing,shridhar2023distilling,zhu2024improving} leverage generative reasoning paths as a supervisory signal to train smaller task-specific models. However, these works solely rely on LLM-generated CoT as the supervisory signal, without considering the capability of PoT. 
~\citet{zhu2023pad} explored the feasibility of PoT as a supervisory signal, but the smaller model used for mathematical reasoning tasks was only trained on the limited dataset, which was insufficient in the experiment and ignored the advantages of CoT in the LLMs. Our proposed MD emphasizes the importance of PoT with CoT as supervisory signals and the smaller model can be effectively activated in multi-path reasoning via self-consistency voting\cite{wang2022self} during inference.

\section{Approach}
We propose a novel distillation framework, Mixed Distillation (MD), to improve the reasoning abilities of smaller models. We achieve this by distilling the CoT and PoT capabilities from LLMs into smaller models. Our overall framework is shown in Figure \ref{framework}. Our paradigm consists of the following steps:

    First, we initiate the process with an LLM and unlabeled datasets, utilizing multiple prompting techniques to elicit the thoughts extraction, denoted as CoT in natural language and PoT in the context of Python program language. Then we filter out incorrectly generated thought paths.

    second, we leverage these mixed thought paths, both CoT and PoT, to train smaller task-specific models. Intuitively, CoT stimulates the internal natural language text knowledge of smaller models, while PoT helps to generate formal intermediate steps, enhancing its reasoning capabilities. Meanwhile, we assume that multi-task learning helps the smaller model to better grasp the distribution of two types of formal data, thus potentially enhancing its multi-path reasoning ability.

    Finally, the smaller model employs these dual capabilities to facilitate multi-path reasoning, deriving the final answer via self-consistent voting during inference.

\subsection{Thoughts Extraction from LLM}

Recent work has observed that LLMs exhibit the capability to generate intermediate reasoning steps\cite{kojima2022large,wei2022chain} and concentrate on distilling its reasoning abilities into smaller models~\cite{buciluǎ2006model, ba2014deep,hinton2015distilling,beyer2022knowledge,fu2023specializing}. Our focus is on distilling the mixed capabilities of LLM into task-specific models to enhance the performance of smaller models.

We employ CoT prompts\cite{wei2022chain} and PoT prompts\cite{chen2022program} to elicit and extract the reasoning thought process of LLMs. As shown in Figure \ref{few-shot}, given an unlabeled dataset, $x_i \in \mathcal{D}$, we begin by devising a prompt template, denoted as $p$, to define how the task should be addressed. Each prompt takes the form of a triplet, $(x_p, r_p, y_p)$, where $x_p$ represents an example input, $y_p$ corresponds to its associated label, and $r_p$ comprises a user-provided reasoning path explaining why $x_p$ can be categorized as $y_p$. We append each input, $x_i$ to the template $p$ and employ it as the input prompt for the LLM to generate reasoning paths and labels as $\hat{r}_i, \hat{y}_i$ for each $x_i\in \mathcal{D}$. Specifically, the PoT few-shot template for StrategyQA, which focuses on commonsense reasoning, is enhanced through the application of CoT as python code annotations\cite{li2023chain} and we show more details in Appendix ~\ref{Appendix_example}. In addition, we will filter out the PoT that Python executors can't execute and the CoT where there is no answer.

\subsection{Mixed Thoughts Distillation}
We first outline the basic paradigm for learning specific task models. Then, we combine CoT and PoT into the training process to extend it. Formally, we represent the dataset as \(D = \{(x_i, y_i)\}_{i=1}^{N}\), where each \(x_i\) is an input, and \(y_i\) is its corresponding output label. In this paper, we focus on text-to-text tasks \cite{raffel2020exploring}.

\paragraph{Standard Specific-task Learning}
The prevalent paradigm for training a task-specific model involves fine-tuning a pre-trained model using supervised data\cite{howard2018universal}. In scenarios where human-annotated labels are unavailable, task-specific distillation\cite{hinton2015distilling,tang2019distilling} employs LLM teachers to produce pseudo-noisy training labels \( \hat{y}_i \) in place of \( y_i \)\cite{wang2021want,smith2022language,arora2022ask}.
For both scenarios, the current model, denoted as $f$, is trained using a paradigm that aims to minimize the loss in label prediction:

\begin{equation}
\label{eq:1}
\mathcal{L} = \frac{1}{N} \sum_{i=1}^{N} \ell(f(x_i), \hat{y}_i)
\end{equation}
where $\ell$ represents the cross-entropy loss between predicted tokens and target tokens. For simplicity and clarity, we use the \( \hat{y}_i \) in Eq.~\ref{eq:1}, representing either human-annotated labels \( y_i \) in the standard fine-tuning scenario or LLM-predicted labels \( \hat{y}_i \) in the context of model distillation.
\begin{table*}
    \centering
    \adjustbox{width=0.91\textwidth}{
        \begin{tabular}{l|cc|cccc}
            \hline
            \multirow{2}{*}{\textbf{Method}} & \multirow{2}{*}{\textbf{\#Params.}} & \multirow{2}{*}{\textbf{\#Training Params}} & \multicolumn{3}{c}{\begin{tabular}[c]{@{}c@{}}\textbf{Mathematical Reasoning}\\ (\%)\end{tabular}} & \begin{tabular}[c]{@{}c@{}} \textbf{Commonsense Reasoning} \\  (\%)\end{tabular} \\ \cline{4-7} 
             &  &  & \textbf{SVAMP} & \textbf{GSM8K} & \textbf{ASDIV} & \textbf{StrategyQA} \\ \hline
                 \rowcolor{gray!20} 
            \multicolumn{7}{c}{\textbf{Closed-Source Models}} \\ 
            GPT-4\cite{openai2023gpt4} & Unknown & 0M & 93.0 & 92.0 & 91.3 & 77.1 \\
            GPT-3.5-Turbo & Unknown & 0M & 82.0 & 77.4 & 75.8 & 71.6 \\  \hline
                \rowcolor{gray!20} 
             \multicolumn{7}{c}{\textbf{Open-Source Models}} \\ 
            LLaMA2\cite{touvron2023llama}\dag & 7B & 7B & 38.0 & 13.3 & 50.7 & - \\
            CodeLlama\cite{roziere2023code}\dag & 7B & 7B & 59.0 & 34.0 & 61.4 & - \\
            WizardMath\cite{luo2023wizardmath}\dag & 7B & 7B & 57.3 & 54.9 & 59.1 & -\\ 
            \hline
                \rowcolor{gray!20} 
            \multicolumn{7}{c}{\textbf{Traditional Distillation}} \\ 
            FlanT5-Large~\cite{fu2023specializing}\dag & 760M & 760M & 6.8 & 6.9 & 10.1 & - \\
            FlanT5-Large + Specialized~\cite{fu2023specializing} & 760M & 760M & 20.4 & 20.2 & 23.8 & - \\ 
            GPT2-Large + Soc~\cite{shridhar2023distilling} & 774M & 774M & - & 21.1 & - & 66.4 \\ 
            GPT-J + Multi-round \& Self-Reflection~\cite{wang2023democratizing} & 6B & - & 55.0 & 33.1 & - & 65.9 \\ 
            T5-large + Distill step by step~\cite{hsieh2023distilling} & 770M & 770M & 65.5 & - & - & - \\ \hline
                \rowcolor{gray!20} 
            \multicolumn{7}{c}{\textbf{Label-Finetuning}} \\ 
            T5-large & 770M & 770M & 7.5 ($\uparrow$0.7) & 7.4 ($\uparrow$0.5) & 11.1 ($\uparrow$1.0) & 50.2 \\
            LLaMA2-7B & 70M & 770M & 50.0 ($\uparrow$12.0) & 10.6
            ($\downarrow$2.7)& 37.3 ($\downarrow$13.4)
            & 51.2 \\
            CodeLlama-7B & 70M & 770M & 39.0 ($\downarrow$20.0) & 9.4 ($\downarrow$24.6) & 24.8 ($\downarrow$36.6) & 50.5 \\ \hline
           \rowcolor{gray!20} 
\multicolumn{7}{c}{\textbf{Single-Path Distillation}} \\
T5-large + CoT & 770M & 770M & 32.5 ($\uparrow$25.7) & 10.5 ($\uparrow$3.6) & 18.9 ($\uparrow$8.8) & 54.3 \\
LLaMA2-7B + CoT & 7B & 160M & 69.5 ($\uparrow$31.5) & 40.1 ($\uparrow$26.8) & 62.2 ($\uparrow$11.5) & 67.7 \\
CodeLlama-7B + CoT & 7B & 160M & 71.0 ($\uparrow$12.0) & 34.2 ($\uparrow$0.2) & 60.0 ($\downarrow$1.4) & 66.4 \\ \hline
            T5-large + PoT & 770M & 770M & 68.0 ($\uparrow$61.2) & 22.5 ($\uparrow$15.6) & 58.1 ($\uparrow$48.0) & 57.3 \\
            LLaMA2-7B + PoT & 7B & 160M & 77.0 ( $\uparrow$39.0 ) & 46.5 ($\uparrow$33.2) & 65.6 ($\uparrow$14.9) & 68.0 \\
            CodeLlama-7B + PoT & 7B & 160M & 83.0 ($\uparrow$24.0) & 51.9 ($\uparrow$17.9) & 67.5 ($\uparrow$6.1) & 67.6 \\ \hline   
            T5-Large + CoT w/ PoT\textsuperscript{*} & 770M*2 & 770M*2 & 70.5 ($\uparrow$63.7) & 24.8 ($\uparrow$17.9) & 57.8 ($\uparrow$47.7) & 58.1 \\
            LLaMA2-7B + CoT w/ PoT\textsuperscript{*} & 7B*2 & 160M*2 & 81.0 ($\uparrow$43.0) & 49.7 ($\uparrow$36.4) & 69.9 ($\uparrow$19.2) & 68.1  \\
            CodeLlama-7B + CoT w/ PoT\textsuperscript{*} & 7B*2 & 160M*2 & 82.5 ($\uparrow$23.5) & 52.0 ($\uparrow$18.0) & 70.2 ($\uparrow$8.8) & 66.7  \\ \hline
                \rowcolor{gray!20} 
            \multicolumn{7}{c}{\textbf{Mixed Distillation (Ours)}} \\ 
           \begin{tabular}[l]{@{}l@{}}T5-Large-MD\\ \end{tabular} &  &  &  &  &  & \\                           
+ CoT & 770M & 770M & 34.5 ($\uparrow$27.7) & 10.6 ($\uparrow$3.7) & 19.0 ($\uparrow$8.9) & 54.3 \\
+ PoT & 770M & 770M & 74.0 ($\uparrow$67.2) & 23.6 ($\uparrow$16.7) & 58.2 ($\uparrow$48.1) & 56.7  \\
+ CoT w/ PoT\textsuperscript{*} & 770M & 770M & 76.0 ($\uparrow$69.2) & 24.6 ($\uparrow$17.7) & 58.3 ($\uparrow$48.2) & 59.1 \\ \hline

\begin{tabular}[l]{@{}l@{}}LLaMA2-7B-MD\\                                      \end{tabular} &  &  &  &  &  & \\   
+ CoT & 7B & 160M & 70.0 ($\uparrow$32.0) & 41.5 ($\uparrow$28.2) & 64.2 ($\uparrow$13.5) & 67.4  \\
+ PoT & 7B & 160M & 80.5 ($\uparrow$42.5) & 51.6 ($\uparrow$38.3) & 66.5 ($\uparrow$15.8) &  66.4 \\
+ CoT w/ PoT\textsuperscript{*} & 7B & 160M & 84.5 ($\uparrow$46.5) & \textbf{53.8} ($\uparrow$40.5) & 70.2 ($\uparrow$19.5) & 69.4 \\ \hline
\begin{tabular}[l]{@{}l@{}}CodeLlama-7B-MD\\                                    \end{tabular}  &  &  &  &  &  & \\   
+ CoT & 7B & 160M & 73.0 ($\uparrow$14.0) & 35.3 ($\uparrow$1.3) & 60.6 ($\downarrow$0.8) &  66.1 \\ 
+ PoT & 7B & 160M & 85.0 ($\uparrow$26.0) & 52.4 ($\uparrow$18.4) & 71.8 ($\uparrow$10.4) & 66.6 \\
+ CoT w/ PoT\textsuperscript{*}& 7B & 160M & \textbf{85.5} ($\uparrow$26.5) & 53.2 ($\uparrow$19.2) & \textbf{73.5} ($\uparrow$12.1) &  \textbf{70.3} \\ \hline

\end{tabular}
    }
    \caption{Accuracy (\%) across tasks:\dag Results are from  \cite{zhu2024improving}. ``+ CoT'' indicates inference via CoT. ``*'' denotes improved performance in distillation using CoT and PoT to generate 10 reasoning paths respectively.
 }
    \label{Table:total_result}
\end{table*}
\paragraph{Muilti-task Learning with CoT and PoT}
In our framework, we use CoT and PoT reasoning paths as combined supervisory signals within the framework of multi-task learning. Specifically, we trained a model $f(x_i) \rightarrow (\hat{r}_{\text{CoT}, i}, \hat{r}_{\text{PoT}, i})$, which can not only generate natural language reasoning path but also provide relevant code reasoning path.
Our overall loss function $\mathcal{L}$ consists of two components:
\begin{equation}
\mathcal{L} = (1 - \lambda) \mathcal{L}_{\text{path\_CoT}} + \lambda \mathcal{L}_{\text{path\_PoT}}
\end{equation}
where $\lambda$ is a weight parameter. We set the default value to 0.5 for a balance between CoT and PoT. Here, 
$\mathcal{L}_{\text{path}}$ is the loss for generating CoT or PoT reasoning paths and predicting labels, defined as:
\begin{equation}
\mathcal{L}_{\text{path}} = \frac{1}{N} \sum_{i=1}^{N} \ell(f(x_i), \hat{r}_i+\hat{y}_i)
\end{equation}
The $\hat{r}_i$ represents the reasoning paths generated by LLM with CoT or PoT, and their respective objective functions are defined as $\mathcal{L}_{\text{path\_CoT}}$ and $\mathcal{L}_{\text{path\_PoT}}$. This is the MD we emphasize. In the
input $x_i$ outlined above, we introduced the concept of task prompts embedded into input examples to train a smaller model for producing distinct reasoning paths. More specifically, we employed \texttt{Let's think step by step} and \texttt{Let's break down the code step by step} to guide the generation of CoT and PoT, respectively. 

Once both CoT and PoT abilities are in the smaller model, multi-path reasoning can be employed via self-consistency voting~\cite{wang2022self} as shown in Figure \ref{framework}. In particular, during inference for the smaller model, input \(x_i\) is concatenated with the guiding prompt phrase \texttt{Let's think step by step} to elicit natural language reasoning paths. The answer result is a final answer list, \(A_{\text{CoT}} = \{a_1, a_2, \ldots, a_n\}\), obtained via \(n\)  n iterations of sampling. Concurrently, by adopting the phrase \texttt{Let's break down the code step by step} similar to the above process, we extract the intermediate code reasoning path. Then utilizing the Python executor, the answer list \(A_{\text{PoT}} = \{b_1, b_2, \ldots, b_n\}\) is acquired. The final prediction of the smaller model, \(\mathcal{P}_{\text{final}}\), is expressed as:
\begin{equation}
\mathcal{P}_{\text{final}} = V(\text{concat}(A_{\text{CoT}}, A_{\text{PoT}}))
\end{equation}
where \(V(\cdot)\) represents a voting function that selects the most frequently occurring answer from the concatenated list of \(A_{\text{CoT}}\) and \(A_{\text{PoT}}\). The \(\text{concat}(\cdot)\) function represents the concatenation of the two lists. This step-by-step thought process along two independent paths ensures that the final prediction is determined through a voting mechanism on the answers procured from each path\cite{wang2022self}.

\section{Experiments}
In this section, we first prove that PoT, as a supervisory signal, outperforms the CoT in enhancing the smaller model's reasoning capabilities in specific tasks(Sec. \ref{pot distillation}). Moreover, our findings emphasize the benefits of Mixed Distillation, which enhanced smaller models' capabilities of single-path reasoning and multiple-path reasoning (Sec. \ref{mm}). We present a comprehensive overview of the experiments based on LLaMA2-7B, CodeLlama-7B, and T5-large. Finally, we validate the generalization of MD (Sec. \ref{gen}).
\paragraph{Datasets} Our experiments primarily centers on following datasets:  SVAMP~\cite{patel2021nlp}, GSM8K~\cite{cobbe2021training}, and ASDIV~\cite{miao2021diverse}. We extend our assessment to StrategyQA\cite{geva2021did}, where we evaluate the capability of commonsense reasoning. More dataset details are in Appendix ~\ref{Appendix_dataset}.

\paragraph{Baselines}
We evaluate MD by comparing experiments with Closed-Source models\cite{openai2023gpt4}, Open-Source Models\cite{touvron2023llama,roziere2023code,luo2023wizardmath} ,Traditional Distillation~\cite{fu2023specializing,shridhar2023distilling,wang2023democratizing,hsieh2023distilling}, Label-Finetuning, Single-Path Distillation and Single-Path reasoning. More details can be found in Appendix \ref{Appendix_baselines}.

\paragraph{Setup} 
We utilize a teacher model grounded in GPT-3.5-Turbo~\footnote{https://platform.openai.com/docs/model-index-for-researchers} in the distillation framework. The experiments cover a wide range of student models, including LLaMA2-7B, CodeLlama-7B, and T5-Large. For the efficient fine-tuning of the LLaMA series, we employ the QLORA~\cite{dettmers2023qlora} method.  In the training process, we set the maximum step to 8000. It's noteworthy that these primary experiments can be conducted on a single GPU with a capacity of 48GB. For the implementation, we make use of the capabilities of HuggingFace Transformers, PyTorch, and Accelerate while adhering rigorously to academic standards.
During the inference process, the number of default total sampling paths is set to 20 in self-consistency voting~\cite{wang2022self}. 
\begin{table*}
    \centering
\scriptsize         \adjustbox{width=0.8\textwidth}{
    \begin{tabular}{l|cc|cc}
    \hline
    \multirow{2}{*}{\textbf{Method}} & \multirow{2}{*}{\textbf{\#Params.}} & \multirow{2}{*}{\textbf{\#Training Params.}} & \multicolumn{2}{c}{\textbf{Mathematical Reasoning}} \\ 
    &  &  & \textbf{GSM8K} & \textbf{ASDIV} \\ \hline
    \rowcolor{gray!20}
    \multicolumn{5}{c}{\textbf{Closed-Source Models}} \\
    GPT-3.5-Turbo  & Unknown & 0M & 77.4 & 75.8 \\ \hline
    \rowcolor{gray!20}
    \multicolumn{5}{c}{\textbf{Open-Source Models}} \\
    LLaMA2\cite{touvron2023llama}\dag & 7B & 7B & 13.3 & 50.7 \\
    CodeLlama\cite{roziere2023code}\dag & 7B & 7B & 34.0 & 61.4  \\
    WizardMath\cite{luo2023wizardmath}\dag & 7B & 7B & 54.9 & 59.1\\ 
    \rowcolor{gray!20}
    \multicolumn{5}{c}{\textbf{Single-Path Distillation}} \\

    LLaMA2-7B + CoT  & 7B & 160M & 19.0 ($\uparrow$5.7) & 50.9 ($\uparrow$0.2) \\ 
    LLaMA2-7B + PoT  & 7B & 160M & 30.0 ($\uparrow$16.7) & 61.2 ($\uparrow$11.5)\\
    LLaMA-7B + CoT w/ PoT\textsuperscript{*} & 7B*2 & 160M*2 & 30.3 ($\uparrow$17.0) & 61.4 ($\uparrow$11.7) \\ \hline
    \rowcolor{gray!20}
    \multicolumn{5}{c}{\textbf{Mixed Distillation (Ours)}} \\
    LLaMA2-7B-MD &  &  & &  \\
    + CoT & 7B & 160M & 20.3 ($\uparrow$7.0) & 52.8 ($\uparrow$2.1) \\
    + PoT & 7B & 160M & 30.5 ($\uparrow$17.2) & 61.6 ($\uparrow$11.9) \\
    + CoT w/ PoT\textsuperscript{*} & 7B & 160M & \textbf{31.2} ($\uparrow$17.9)& \textbf{62.6} ($\uparrow$12.9) \\ \hline
    \end{tabular}}
    \caption{Accuracy (\%) across tasks which demonstrate the generalizability of Mixed Distillation.}
    \label{Table:generalizability}
\end{table*}

\subsection{PoT Distillation Enhanced Reasoning}
\label{pot distillation}
In this subsection, we present the effectiveness of PoT on specialized datasets. By extracting supervisory signals from LLM and distilling these single-format signals into smaller models, we then assess their performance on the respective test sets.

Table \ref{Table:total_result} shows the experimental results, which prove that various models employing the PoT distillation outperform those utilizing the CoT distillation and Label-Finetuning in mathematics and common sense reasoning tasks. For example, T5-Large exhibits a notable improvement of 61.2\% on SVAMP. Similarly, LLaMA2 shows enhancements of 33.2\% on GSM8K and 14.9\% on ASDIV. Meanwhile, T5-Large, LLaMA2, and CodeLlama achieve gains of 7.1\%, 17.8\%, and 17.4\% respectively, compared to Label-Finetuning On StrategyQA. In addition, we observe that compared with T5-Large model, LLaMA models with less training parameters but larger fixed parameters showed excellent performance. In particular, CodeLlama notably achieves 82.5\% accuracy on the SVAMP task, marking a 23.5 \% improvement.
\begin{figure*}[!htbp]
    \centering
    \includegraphics[width=1\linewidth]{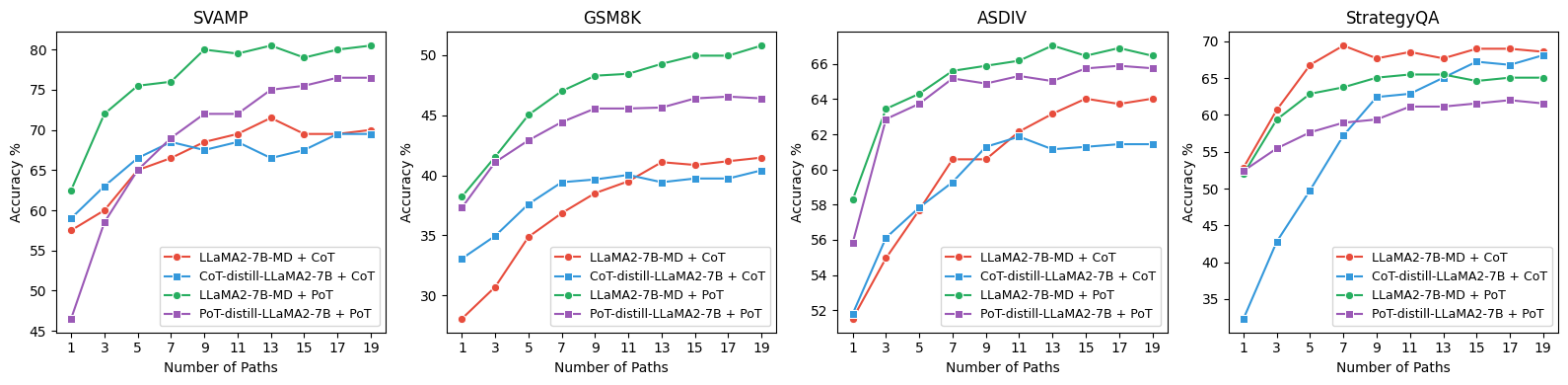} 
    \caption{Performance comparison with Mixed Distillation and Single-Path Distillation on SVAMP, GSM8K, ASDIV, and StrategyQA based LLaMA2-7B.}
    \label{combined_Llama}
\end{figure*}
 \subsection{Mixed Distillation Enhanced Reasoning}
 \label{mm}
In this subsection, we show the effectiveness of MD on specialized datasets from two aspects: enhancing single-path reasoning and multi-path reasoning. Following the extraction of supervisory signals from LLM and the distillation of these mixed-format signals into smaller models, we evaluate their performance on corresponding test sets.

\begin{figure}
    \centering \includegraphics[width=0.91\linewidth]{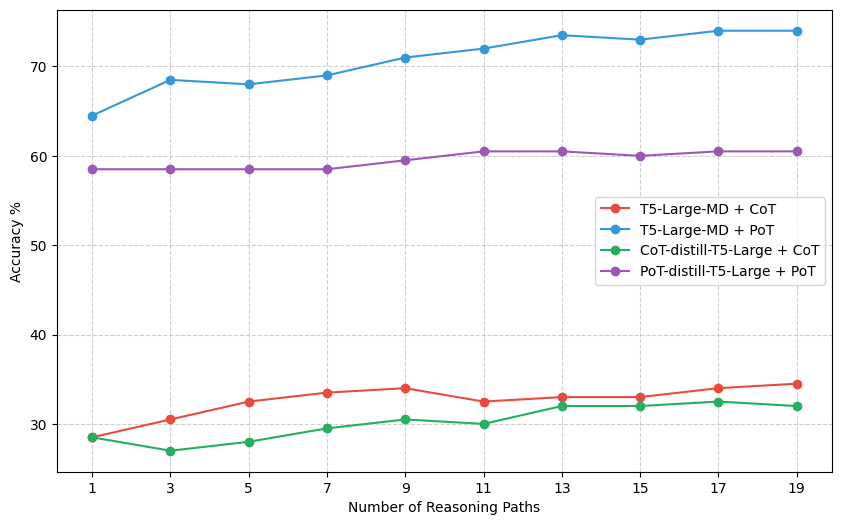}
    \caption{Performance comparison with Mixed Distillation and Single-Path Distillation on SVAMP based T5-Large.}
    \label{t5}
\end{figure}

\begin{figure*}[!htbp]
    \centering
    \includegraphics[width=1.0\linewidth]{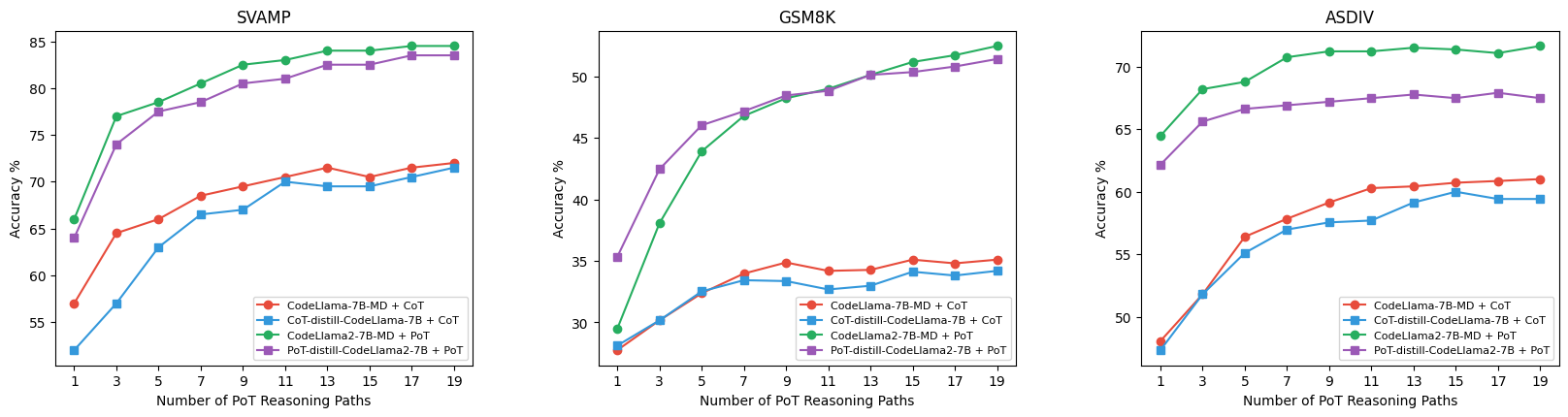} 
    \caption{Performance comparison with Mixed Distillation and Single-Path Distillation on SVAMP, GSM8K, ASDIV based CodeLlama-7B.}
    \label{combined_CodeLlama}
\end{figure*}

\paragraph{Mixed Distillation Enhances Single-Path Reasoning}
As shown in Table \ref{Table:total_result}, the CoT and PoT abilities of models are improved by mixed distillation. For example, compared with CoT distillation on ASDIV, the CoT ability of LLaMA2 exhibits a 2\% increase. Similarly, T5-Large's PoT capability shows a 6.0\% improvement over PoT distillation on SVAMP. Specifically, Figure \ref{combined_Llama} displays the capabilities of LLaMA2 across different datasets. It is worth noting that with the increase of the number of sampling inference paths, the PoT ability of the model using MD is always better than that of the model trained by Single-Distillation, and the difference observed in the sampling interval of 10-13 paths is the most significant. In addition, CoT capability with MD exceeds the performance of smaller models with single distillation as the number of sampling paths exceeds 13. Figure \ref{combined_CodeLlama} shows the performance of CodeLlama. MD has an obvious advantage on ASDIV, showing a 4.3\% increase in PoT capability when using 20 sampling paths, compared to using PoT distillation. Figure \ref{t5} also shows the performance of T5-Large to validate that MD enhances single-path reasoning. 
\paragraph{Mixed Distillation Enhances Multi-Path Reasoning} Using
``+CoT w/PoT'' for multi-path reasoning during inference, various models achieve state-of-the-art performances across different tasks. Notably, LLaMA2 excels on GSM8k, achieving an accuracy of 53.8\%, which marks an impressive improvement of 40.5\%. Similarly, CodeLlama shows remarkable results on the SVAMP and ASDIV tasks, reaching accuracies of 85.5\% and 73.5\%, respectively, and registering boosts of 26.5\% and 12.1\%. Additionally, T5-Large stands out in the StrategyQA task with an accuracy of 59.1\%, indicating a 7.9\% increase over the Label-Finetuning. Furthermore, a single model from MD outperforms two individually distilled models, with LLaMA2 gaining 3.5\% and 4.1\% improvement on SVAMP and GSM8K, respectively.

\subsection{Generalization}
\label{gen}
In the above experiments, we have proved the effectiveness of MD in the generalization of models and tasks. Furthermore, we conduct experiments to validate the framework's generalization across varying training set sizes and out-of-distribution scenarios, based on LLaMA2. For the training set sizes generalization assessment, we perform distillation using proportions of 20\%, 40\%, 60\%, and 80\%, on SVAMP. To evaluate the ability of out-of-distribution generalization, we evaluate the model trained on SAVMP for GSM8K and ASDIV datasets.

\begin{figure}
    \centering \includegraphics[width=1.0\linewidth]{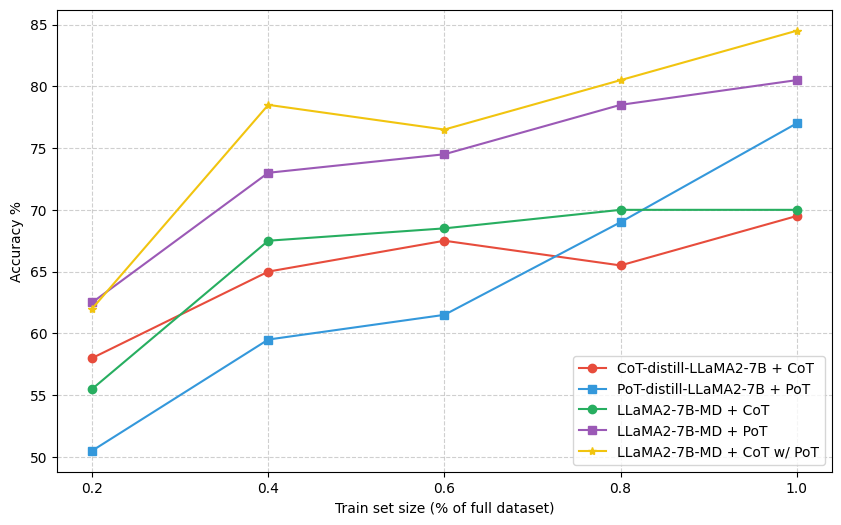}
    \caption{Performance comparison with different train set size on SVAMP based LLaMA2-7B.}
    \label{trainset}
\end{figure}

\subsubsection{Training Set Size}
As shown in Figure \ref{trainset}, with the increase of data volume, the performance of models generally improves. Specifically, when the dataset size surpasses 75\%, LLaMA2 using PoT as the supervisory signal outperforms that with CoT. This observation shows that it takes a certain dataset size to learn the PoT capability. 
In the case of Mixed Distillation, across the dataset size range of 20\% to 100\%, MD always enhances PoT's capability in comparison to PoT Distillation. Moreover, when the dataset size exceeds 40\%, compared with CoT distillation, the learning ability of CoT is improved with a 2\% improvement.
By incorporating multi-path reasoning during inference, the model achieves its optimal performance, providing evidence for the effectiveness of multi-path reasoning in Mixed Distillation.

\begin{figure}
    \centering \includegraphics[width=1.0\linewidth]{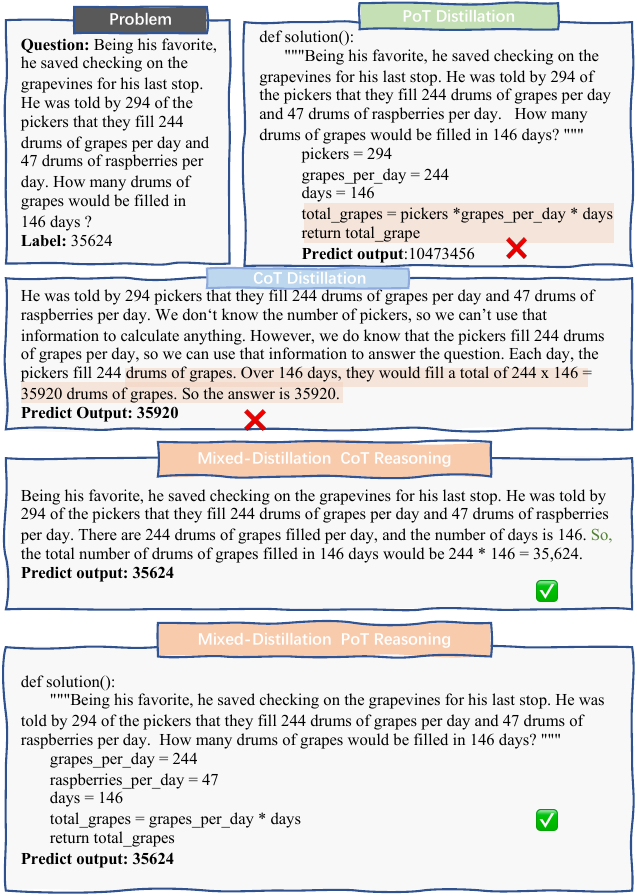}
    \caption{Case study of different distillation methods based LLaMA2-7B.}
    \label{figure:case1}
\end{figure}

\subsubsection{Out-of-Distribution Evaluation}
As shown in Table \ref{Table:generalizability}, on GSM8K and ASDIV, models using MD via CoT reasoning showed improvements of 1.29\% and 1.88\% over standard CoT distillation. Similarly, MD enhances the performance of smaller models compared to PoT distillation. Moreover, the model using MD with multi-path reasoning during inference leads the model to achieve optimal performance, attaining accuracies of 31.2\% and 62.6\% on GSM8K and ASDIV, with substantial improvements of 17.9\% and 12.9\%, respectively.

\section{Case Study}
To comprehensively elucidate the effectiveness of Mixed Distillation, we present the actual output of smaller models for reasoning tasks, as shown in Figure \ref{figure:case1}. 
For the question labeled 35624, it is obvious that the correct answers are obtained using MD involving CoT or PoT. However, when using CoT or PoT distillation, there will be errors in the reasoning process. Specifically, the error in CoT distillation is attributed to the inability to effectively compute 244*146, a common issue indicative of poor computational capability in CoT. We show more details in Appendix \ref{case_smaller}.

\section{Conclusion}
In this paper, we first spotlight the PoT as an effective supervisory signal for model distillation to enhance smaller models' reasoning abilities. Then we introduce a novel framework, Mixed Distillation, that distills the reasoning paths of CoT and PoT from LLM into smaller models. Our experimental results demonstrate that the MD enhances the smaller models' single-path reasoning and multi-path reasoning, which is not only better than the two individually distilled models but also surpasses LLMs on SVAMP. Comparative analysis and experimental results show that our MD boosts reasoning can effectively extract two different forms of capabilities, CoT and Pot from LLM, to improve the reasoning ability of smaller models.

\section*{Limitations}
Our work has proven that the MD technology can improve the reasoning ability of small models. However, this technique has several limitations. Firstly, our findings focus on reasoning tasks in English and have not been verified in a multilingual setting. Secondly, MD relies on the closed model, GPT-3.5-turbo, which may raise the potential for biases. Thirdly, our technology uses the generated intermediate reasoning steps to predict the final result, and the direct relationship between these intermediate reasoning steps and the final answer is still unproven. Caution should be taken when displaying MD to users.

\bibliography{anthology,custom}
\clearpage
\appendix
\section{Datasets}
\label{Appendix_dataset}
We provide detailed information about the datasets, including their sources and the initial release of the authors in the experiments.
\begin{itemize}
    \item \textbf{SVAMP:} The dataset was originally released in \cite{patel2021nlp} and made publicly available at \url{https://github.com/arkilpatel/SVAMP}. We obtained the dataset from \url{https://huggingface.co/datasets/ChilleD/SVAMP}.
    \item \textbf{GSM8K:} The dataset was originally released in \cite{cobbe2021training} and made publicly available at \url{https://github.com/openai/grade-school-math}. We obtained the dataset from \url{https://huggingface.co/datasets/gsm8k}. 
    \item \textbf{ASDIV:} The dataset was originally released in \cite{miao2021diverse} and made publicly available at \url{https://github.com/chaochun/nlu-asdiv-dataset}. We obtained the dataset from \url{https://github.com/chaochun/nlu-asdiv-dataset/blob/master/dataset/ASDiv.xml}.
    \item \textbf{StrategyQA:} The dataset was originally released in \cite{geva2021did} and made publicly available at \url{https://github.com/eladsegal/strategyqa}. We obtained the dataset from \url{https://github.com/eladsegal/strategyqa/tree/main/data/strategyqa}.
\end{itemize}

 For ASDIV, we randomly selected 695 instances for the test set based on the question grade distribution in the training set. For StrategyQA, we use the dev set as the test set. The statistical information for the datasets is available in Table \ref{Table:dataset}.

\begin{table*}
    \centering
    \adjustbox{width=1.0\textwidth}{
\begin{tabular}{lccc}
\hline
Dataset & Train set size & \multicolumn{1}{l}{Test set size} & Example \\ \hline
SVAMP~\cite{cobbe2021training} & 700 & 300 & \begin{tabular}[c]{@{}c@{}}Paige was helping her mom plant flowers and together they planted some seeds. \\ They put 10 seeds in each flower bed. If there are 45 flowerbeds\\ How many seeds did they plant?\end{tabular} \\ \hline
GSM8K~\cite{cobbe2021training} & 7473 & 1319 & \begin{tabular}[c]{@{}c@{}}Janet\textbackslash{}u2019s ducks lay 16 eggs per day. She eats three for breakfast every morning \\ and bakes muffins for her friends every day with four.  She sells the remainder at \\ the farmers' market daily for \$2 per fresh duck egg. \\ How much in dollars does she make every day at the farmers' market?\end{tabular} \\ \hline
ASDIV~\cite{miao2021diverse} & 1610 & 695 & \begin{tabular}[c]{@{}c@{}}Edward spent $13. Now he has $6.\\ How much did Edward have before he spent his money?\end{tabular} \\ \hline
StrategyQA~\cite{geva2021did} & 2061 & 229 & \begin{tabular}[c]{@{}c@{}}Will the Albany in Georgia reach a\\  hundre thousand occupants before the one in New York?\end{tabular}  \\ \hline
\end{tabular}
    }
    \caption{Details of dataset, including SVAMP, GSM8K, ASDIV, and StrategyQA.}
    \label{Table:dataset}
\end{table*}
\begin{figure}
    \centering
    \includegraphics[width=\linewidth]{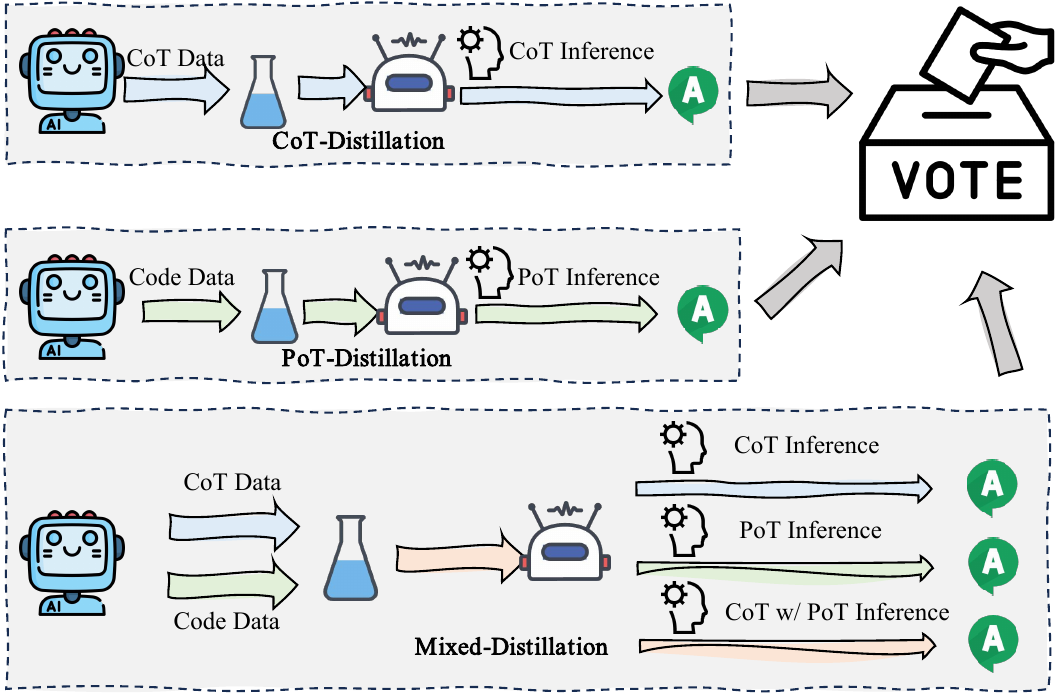}
    \caption{Framework diagram for different distillation methods.}
    \label{dif_distill}
\end{figure}

\begin{figure*}
    \centering
    \includegraphics[width=\linewidth]{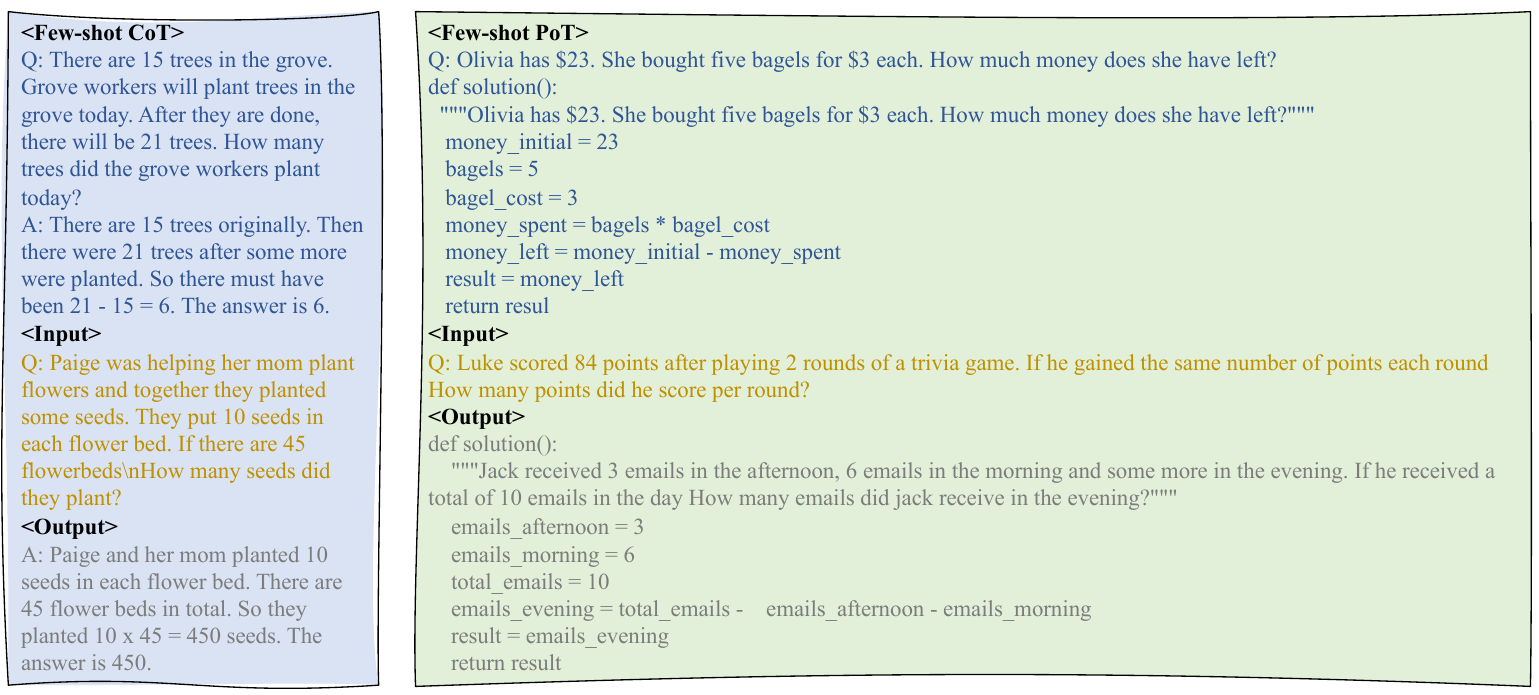}
    \caption{Example of Mixed Distillation: extracting and distilling CoT and PoT from large Language Models to task-specific smaller models on SVAMP, GSM8K,ASDIV.}
    \label{few-shot}
\end{figure*}

\begin{figure*}
    \centering
    \includegraphics[width=\linewidth]{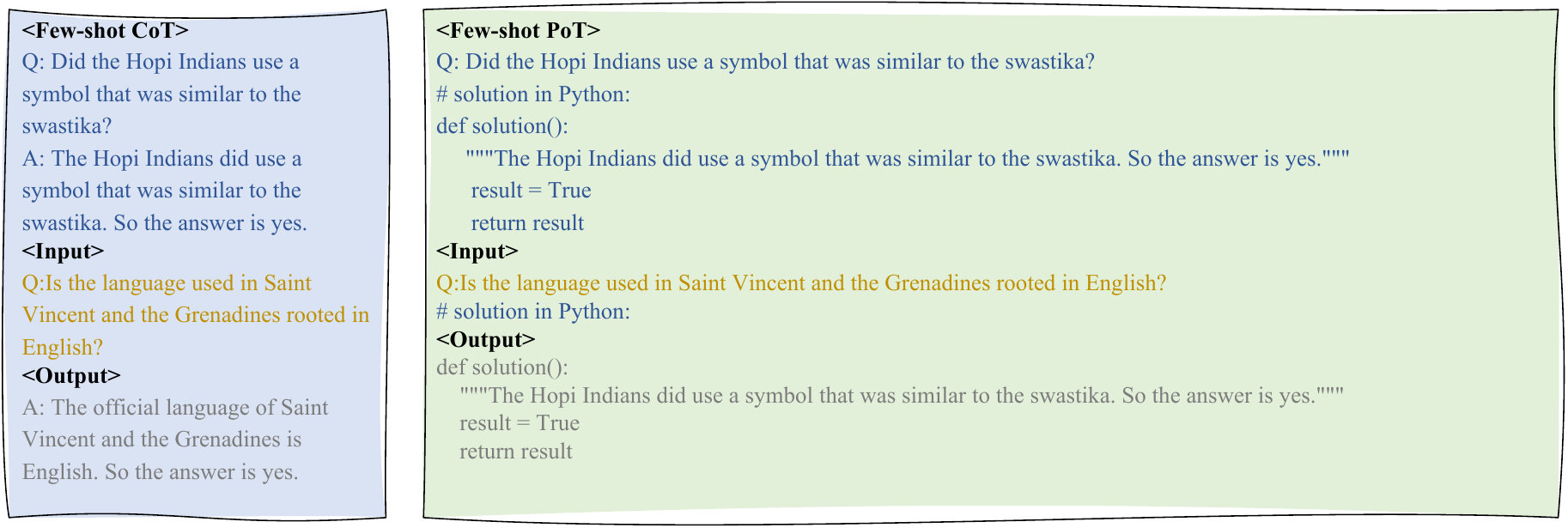}
    \caption{Example of Mixed Distillation: extracting and distilling CoT and PoT from large Language Models to task-specific smaller models on StrategyQA.}
    \label{few-shot_csqa}
\end{figure*}
\section{Prompt Examples}
\label{Appendix_example}
For the datasets SVAMP, GSM8K, and ASDIV, the few-shot prompts are shown in Figure \ref{few-shot}. For StrategyQA, they are displayed in Figure \ref{few-shot_csqa}. We draw inspiration from \cite{li2023chain} and add CoT as annotations.

\section{Baselines}
\label{Appendix_baselines}
In the section, we show more details, including Closed-Source models, Open-Source models, Traditional Distillation, Label-Finetuning, Single-Path Distillation and Single-path Reasoning strategies, aiming at providing a comprehensive comparison between MD and a series of existing methods.

\paragraph{Closed-Source Models}
Advanced Language Models, such as OpenAI's GPT-4\cite{OpenAI2023GPT4TR} and GPT-3.5-Turbo, have achieved state-of-the-art results across various NLP tasks\cite{zhao2023survey,kasneci2023chatgpt,chang2023survey,hao2023reasoning}. Trained on extensive datasets, these models comprehend complex language structures and generate text resembling human expression. Comparing them with closed-source models like GPT-4 is helpful to evaluate the reasoning gap between smaller models with MD and closed-source models.
\paragraph{Open-Source Models}
There are a series of models in the field of open source NLP. Notably, LLaMA2\cite{touvron2023llama}, publicly released by Meta, demonstrates competitiveness and makes a significant contribution to academic research. CodeLlama, an adaptation of LLaMA, excels in diverse reasoning tasks, particularly showcasing proficiency in code-related capabilities\cite{roziere2023code}. WizardMath, fine-tuned based on LLaMA with enhanced evolution instructions, effectively competes in mathematical reasoning tasks\cite{luo2023wizardmath}. 
\paragraph{Traditional Distillation}
Knowledge distillation~\cite{buciluǎ2006model, ba2014deep,hinton2015distilling,beyer2022knowledge,fu2023specializing} has demonstrated the effectiveness in improving smaller models.
\cite{fu2023specializing} distills LLMs' multi-step reasoning into smaller models for better mathematical reasoning. \cite{shridhar2023distilling} improves mathematical skills by distilling LLMs' problem decomposition abilities. \cite{wang2023democratizing} and \cite{hsieh2023distilling} focus on distilling reflective thinking and using LLM-generated CoT as supervisory signals, respectively.

\paragraph{Label-Finetuning}
Label fine-tuning is a supervised learning method, in which the trained model is adjusted to do better on a specific task. It uses a small set of labeled data to adjust the model's settings, which was initially trained on a broad dataset. The main goal is to make the pre-trained model work better in the tasks. We use the training set questions and labels for model training to establish this benchmark.

\paragraph{Single-Path Distillation and Reasoning}
Single-Path Distillation involves distilling smaller models using data in a single format, including CoT-distill model, PoT-distill model and  and a unified ensemble from two individually distilled models.
Single-Path Reasoning refers to selecting only one capability, either CoT or PoT for inference. '+CoT' indicates CoT inference on the task, and '+PoT' indicates PoT inference on the task. As shown in Figure \ref{dif_distill}, the final answer is obtained through voting during inference. 
\section{Case Analysis}
\subsection{Case Analysis in LLM}
\begin{figure*}
    \centering
    \includegraphics[width=1.0\linewidth]{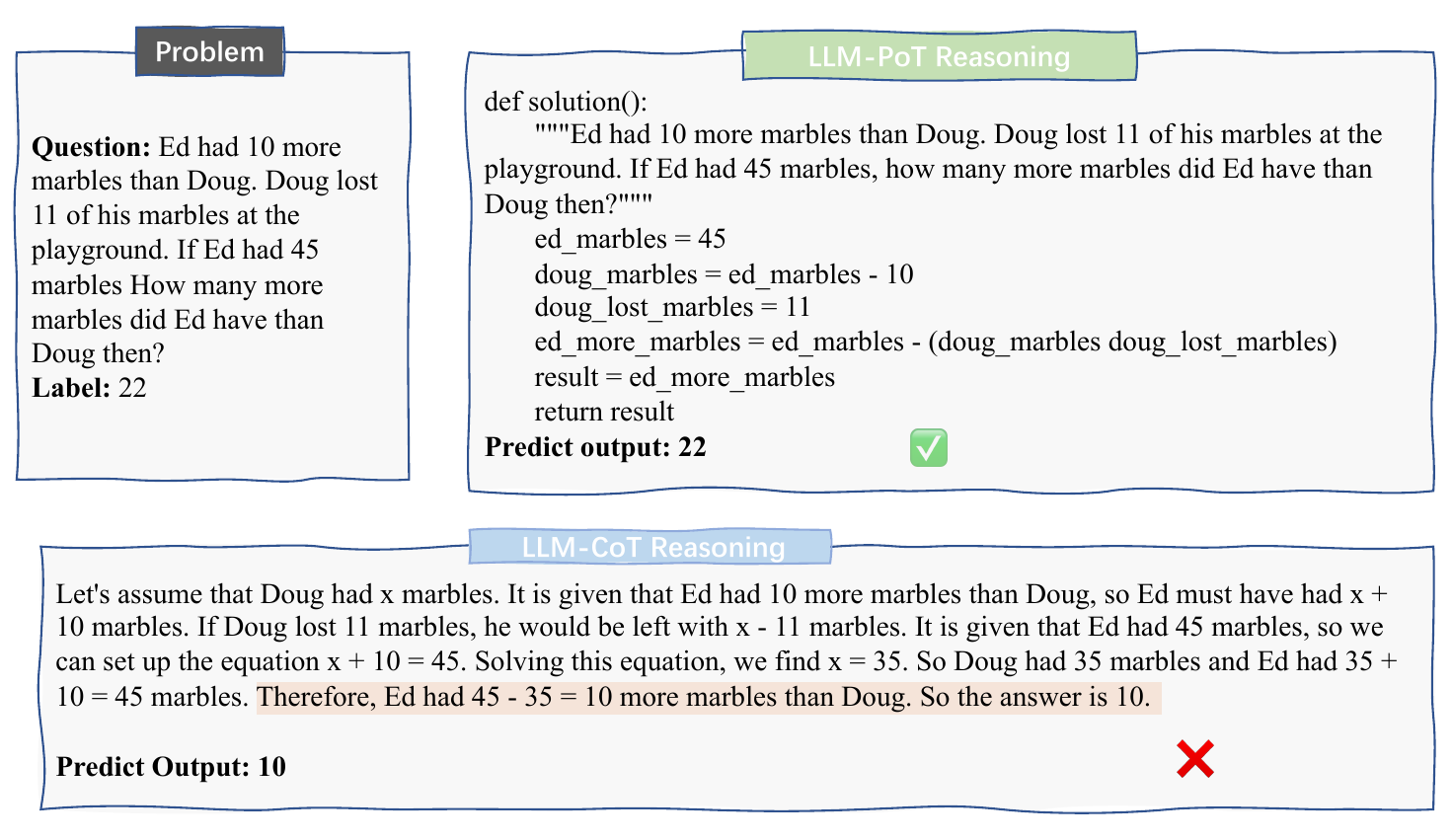}
    \caption{PoT yields the correct solution, whereas CoT falls short in GPT-3.5-Turbo.}
    \label{figure:llm_case1}
\end{figure*}

\begin{figure*}
    \centering
    \includegraphics[width=1.0\linewidth]{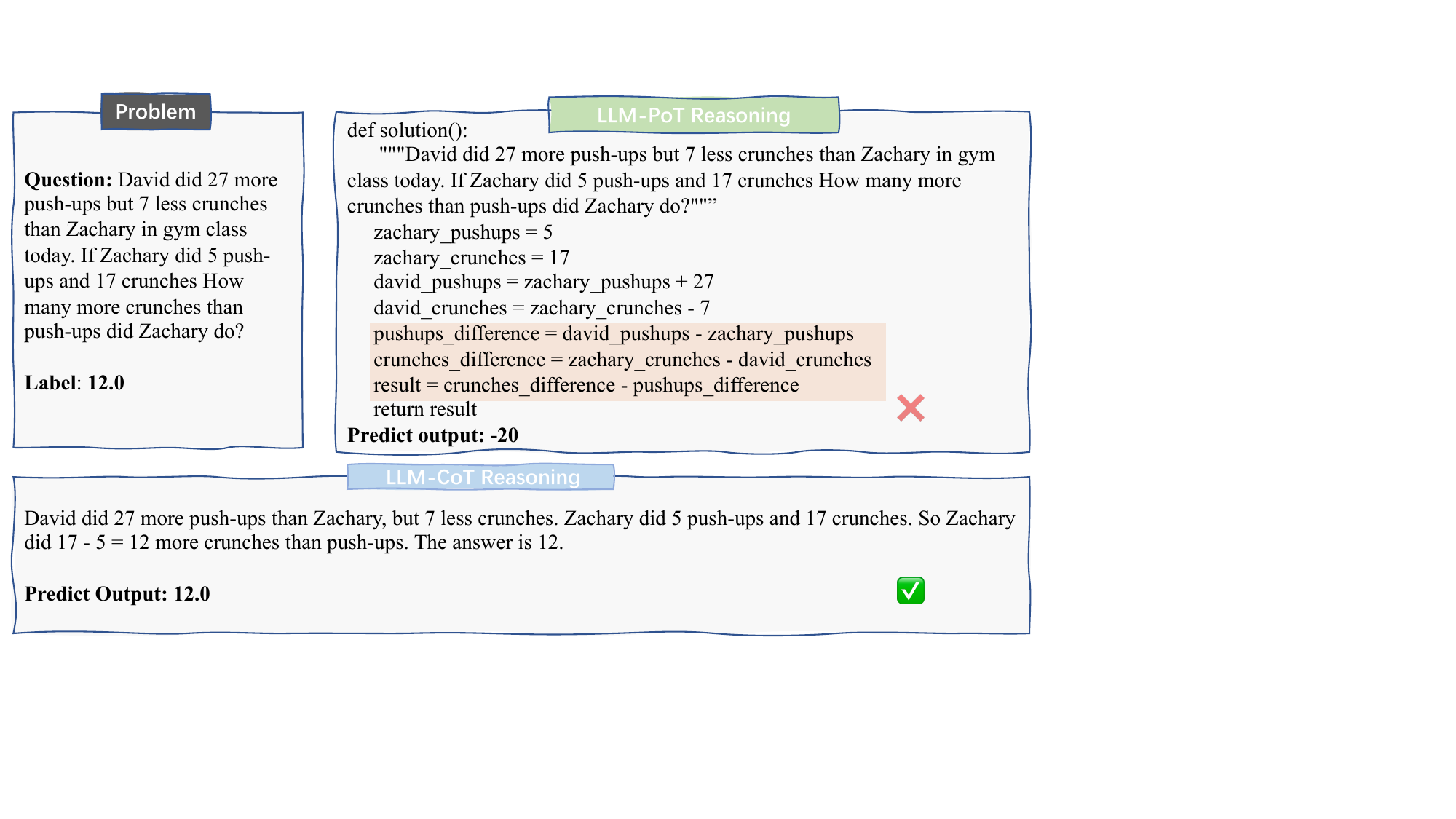}
    \caption{CoT yields the correct solution, whereas PoT falls short in GPT-3.5-Turbo.}
     \label{figure:llm_case2}
\end{figure*}

Despite PoT demonstrating superiority over CoT in LLMs\cite{chen2022program,gao2023pal}, recent work has identified distinct weaknesses for CoT and PoT\cite{yue2023mammoth}. As shown in Figure \ref{figure:llm_case1}, CoT overlooks the statement ``Doug lost 11 of his marbles at the playground,'' leading to a reasoning error. Similarly, in Figure \ref{figure:llm_case2}, PoT misinterprets the question ``How many more crunches than push-ups did Zachary do?'' resulting in a reasoning error. Thus, combining these reasoning skills may compensate for their individual weaknesses.

\label{casellm}

\subsection{Case Analysis in smaller models}
\label{case_smaller}
\begin{figure*}
    \centering
    \includegraphics[width=1.0\linewidth]{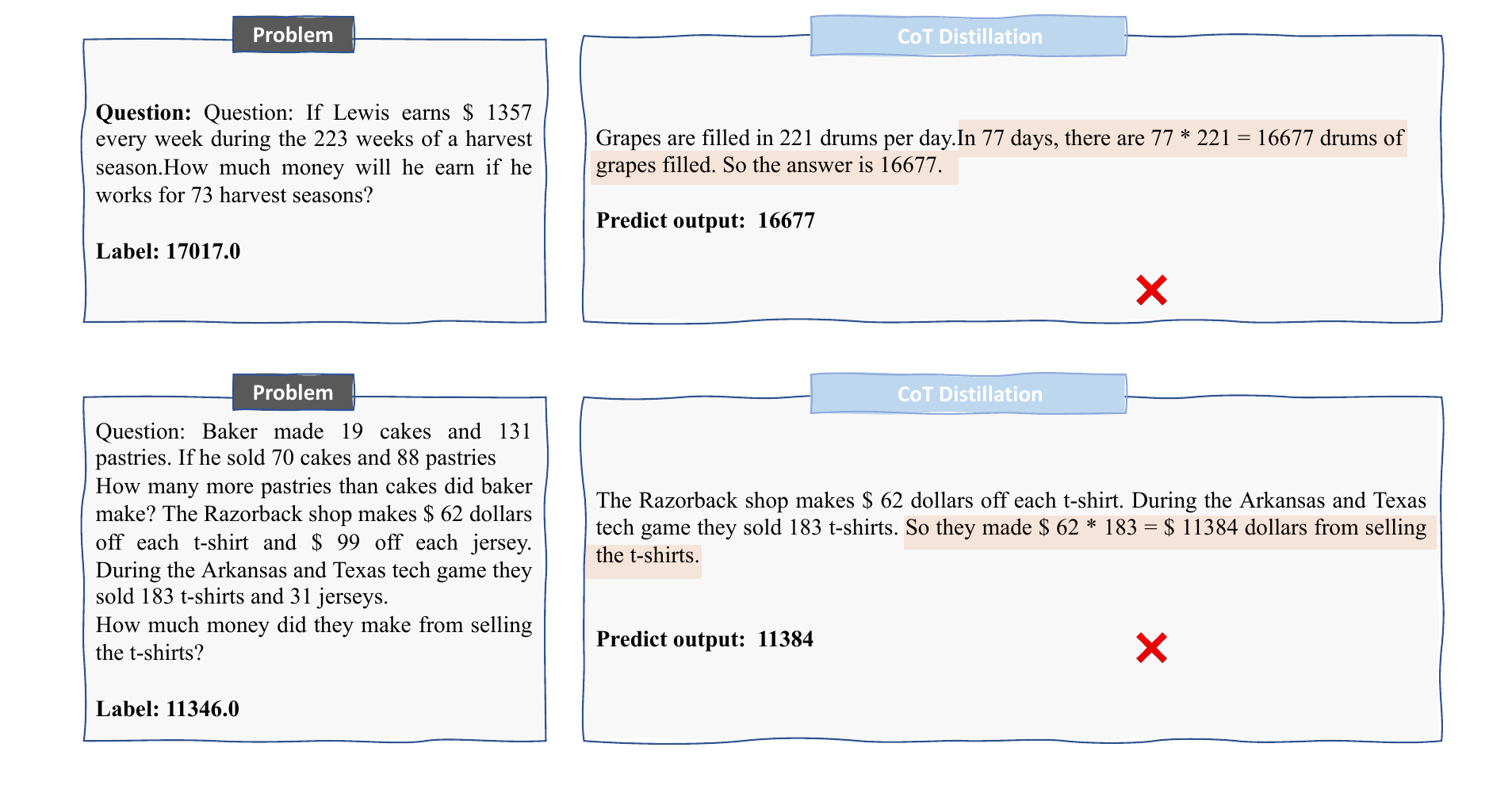}
    \caption{Error cases based LLaMA2-7B in CoT Distillation.}
     \label{figure:small_case1}
\end{figure*}

\begin{figure*}
    \centering
    \includegraphics[width=1.0\linewidth]{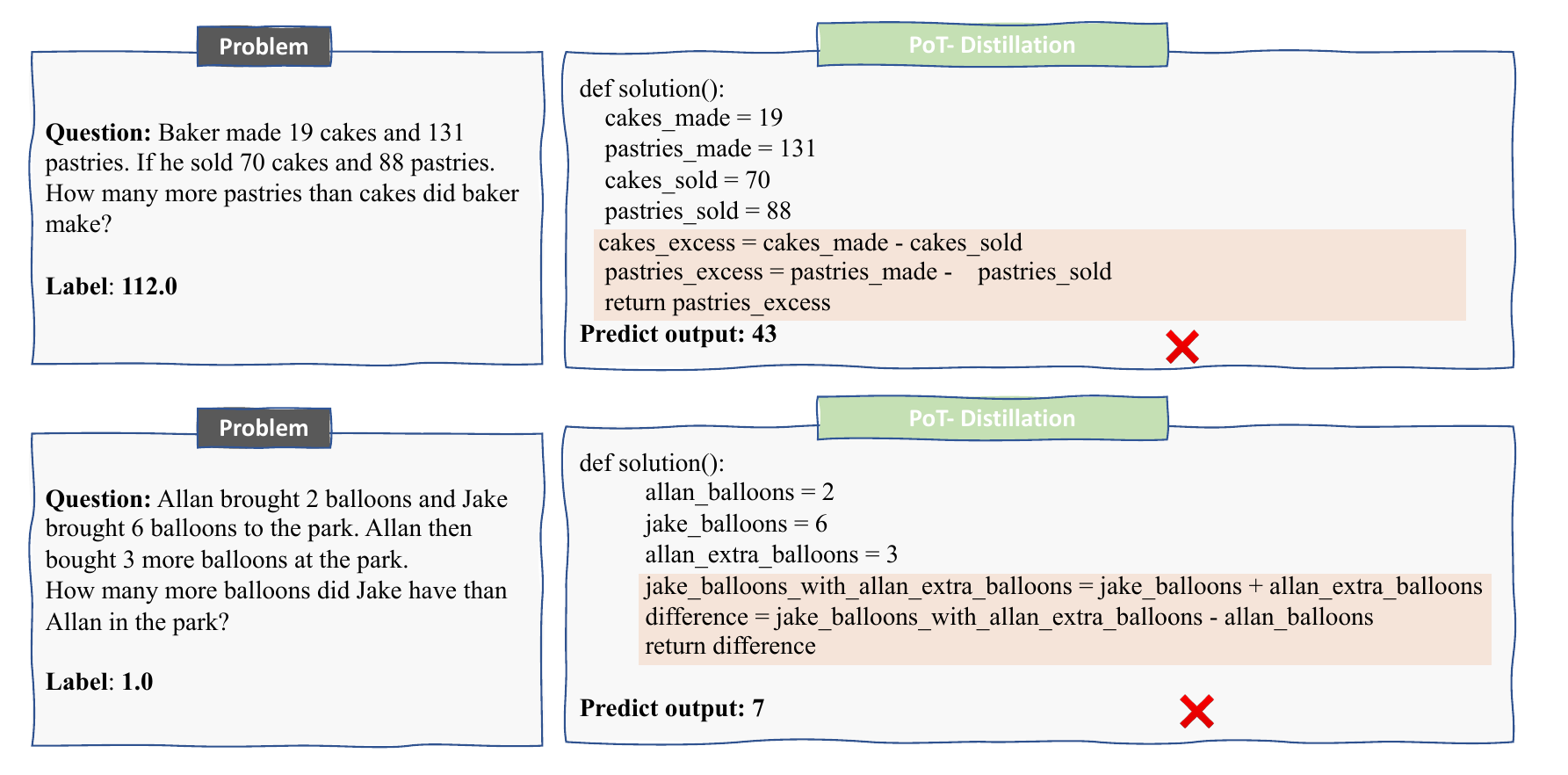}
    \caption{Error cases based LLaMA2-7B in PoT Distillation.}
     \label{figure:small_case2}
\end{figure*}

\begin{figure*}
    \centering
    \includegraphics[width=1.0\linewidth]{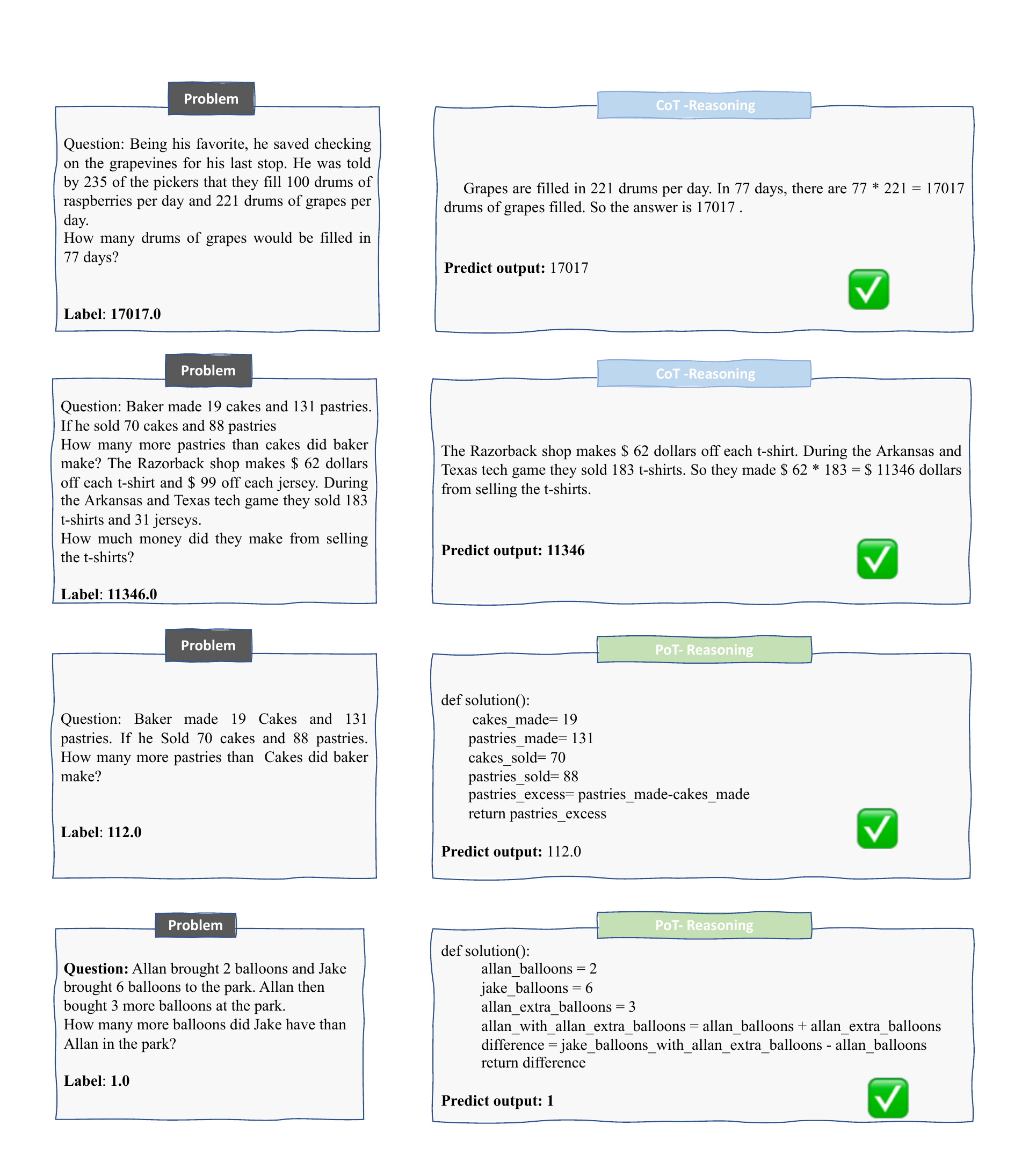}
    \caption{Correct cases based LLaMA2-7B in Mixed Distillation.}
     \label{figure:small_case3}
\end{figure*}

We propose more case studies, focusing on smaller models. As shown in Figure \ref{figure:small_case1}, our experimental results reveal that CoT Distillation encounters challenges in handling complex numbers, such as 77*221 and 62*183. Conversely, PoT Distillation struggles with understanding problems involving multiple terms, such as when irrelevant 
conditions are added, like ``if he sold 70 cakes and 88 pastries'', leading to error reasoning steps, and an inability to understand statements like 'Allan bought 3 more balloons' as shown in Figure \ref{figure:small_case2}. However, MD can effectively deal with these shortcomings, thus improving the results as shown in Figure \ref{figure:small_case3}.

\end{document}